\documentclass[11pt]{article}
\input epsf 

\usepackage{jair}
\usepackage{theapa}

\jairheading{27}{2006}{441-464}{03/06}{12/06}
\ShortHeadings{Set Intersection and Consistency in Constraint Networks}
{Zhang \& Yap}
\firstpageno{441}

\usepackage{graphics}
\usepackage{epstopdf}
\usepackage{amssymb}
\usepackage{latexsym}

\usepackage{pictex}
\def \color#1]#2{}

\newtheorem{theorem} {Theorem}
\newtheorem{example} {Example}

\newtheorem{corollary} {Corollary}
\newtheorem {proposition} {Proposition}
\newtheorem{lemma} {Lemma}
\newtheorem {definition} {Definition}

\newcommand{\figref}[1] {Figure~\ref{#1}}

\newcommand{\ra} {\rightarrow}
\newcommand{\network}[1] {$\mathcal {#1}$}
\newcommand{\set}{\mbox{$\mathcal {S}$}}
\newcommand{\tree}{$\mathcal {T}$} 

\newcommand{\QED} {\mbox{$\Box$}}
\newcommand{\hide}[1]{} 
\newcommand{\treeconvex}{tree convex}
\newcommand{\treeconvexity}{tree convexity}

\begin{document}


\title{Set Intersection and Consistency in Constraint Networks }

\author{\name Yuanlin Zhang \email yzhang@cs.ttu.edu \\
        \addr Department of Computer Science, Texas Tech University \\
	 Lubbock, TX 79414 USA
       \AND
       \name Roland H. C. Yap \email ryap@comp.nus.edu.sg \\
       \addr Department of Computer Science, National University of Singapore \\
	3 Science Drive 2, Singapore 117543 
}


\maketitle

\hide{
plan:

0. abstract: shorter, more on new results
1. Intro: same except the layout paragraph
2. Preliminaries: same 
3. Properties on set intersection: remove old results (lemmas and
   corollaries), contain only tree convex results and set intersection
   results.
4. Set intersection and consistency: same
5. Tree convex constraints: same
6. On tightness of constraints
   a. weakly tight networks (section 6 in the original submission)
   b. making weakly tight networks globally consistent (section 8)
   c. more about properties on weakly tight networks (CP04)
   d. duly adaptive consistency (CP04)
7. Discussion: same 
11 Sep -- 
Jun 8 2006 revision 
}

\begin{abstract} 

In this paper, we show that there is a close relation 
between consistency in a constraint network
and set intersection. A proof schema is provided as a generic way
to obtain consistency properties from
properties on set intersection. This approach 
not only simplifies the understanding of and unifies many existing consistency
results, but also directs the study of consistency to that of
set intersection properties in many situations, as demonstrated by the 
results on the convexity 
and tightness of constraints in this paper. Specifically, we identify a new 
class of {\em tree convex} constraints where local consistency ensures 
global consistency. This generalizes row convex constraints. 
Various consistency results are also obtained on constraint networks
where only {\em some}, in contrast to {\em all} in the existing work,
constraints are tight. 
\end{abstract}
            
            

\section{Introduction}

A constraint network consists of a set of variables over
finite domains and a system of constraints over those variables.
An important task is to find an assignment for all the variables such
that all the constraints in the network are satisfied.
If such an assignment exists, the network is {\em satisfiable} or
{\em globally consistent}, and the assignment is called a solution.
The problem of determining the global consistency of a 
general constraint network is NP-complete.
Usually a search procedure is employed to find a solution.
In practice, due to efficiency considerations, the search is
usually equipped with a filtering algorithm that prunes values of a variable
or the combinations of values of a certain number of variables that cannot
be part of any solution.
The filtering algorithm can make a constraint network
{\em locally consistent} in the sense that a consistent assignment of
some variables can always be extensible to a new variable.
An important and interesting question on local consistency is:
\begin{quote}
Is the local consistency obtained sufficient to determine the global consistency
of the network without further search?
As the filtering algorithm is of polynomial complexity,
a positive answer would mean that the network
can be solved in polynomial time.
\end{quote}

Much work has been done to explore the relationship between local and global
consistency in particular and the properties of local consistency in
general. One direction is to make use of the topological structure
of a constraint network. A classical result is
that when the graph of a constraint network is a tree, 
arc consistency of the network is sufficient to
ensure its global consistency \cite{Fre82}.

The second direction\footnote{
There is a difference between the work concerned here and
that studying the tractability of 
constraint languages \cite<e.g.,>{Sch78,JCG97}. The latter considers the 
problems whose constraints are from a {\em fixed set of relations} 
while the former studies constraint networks with {\em special properties}. 
}
makes use of semantic properties of the constraints.
For {\em monotone constraints}, path consistency
implies global consistency \cite{Mon74}. Van Beek and Dechter
\citeyear{vBD95} generalize monotone constraints to a larger
class of {\em row convex constraints}. Dechter \citeyear{Dec92b}
shows that a certain level of consistency in a constraint network whose
domains are of limited size ensures global consistency. 
Later, Van Beek and Dechter \citeyear{vBD97}
study the consistency of constraint networks with tight and
loose constraints.


The existing work along the two approaches has used specific and different
techniques to study local and global consistency. 
In particular, there is little commonality in the details of the
existing work.
In much of the existing work, the techniques and consequently
the proofs given are developed specifically for the results concerned.

In this paper, we show how much of this work can be connected together through
a new approach to studying consistency in a constraint network.
We unite two seemingly disparate areas: the study of 
{\em set intersection} on special sets and the study of {\em $k$-consistency}
in constraint networks.
In fact, $k$-consistency can be expressed in terms of set intersection,
which allows one to obtain relationships between local and global
consistency in a constraint network through the properties of set
intersection on special sets.
The main result of this approach is a proof schema that can be 
used to lift results from
set intersection, which are rather general, to particular consistency
results on constraint networks.
One benefit of the proof
schema lies in that it provides a modular way to greatly simplify
the understanding and proofs of consistency results. This
benefit is considerable as often the proofs of many existing
results are complex and ``hard-wired''.
Using this new approach, we show that it is precisely the various
properties of set intersection that are the key to those results.
Furthermore, the proofs become mechanical.


The following sketch illustrates briefly the use of our approach.
One property of set intersection is that if the
intersection of every pair ({\em 2}) of {\em tree convex sets}
(see Section~\ref{sec:set-intersection}) is not empty, the
intersection of the whole collection of these sets is not empty too.
From this property, we can see that the local information on
the intersection of every pair of sets gives rise to the global information on
the intersection of all sets. Intuitively, this relationship between
the local and global information corresponds to obtaining global
consistency from local consistency. The proof schema is used to
lift the result on tree convex sets to the following consistency result. 
For a binary constraint network of tree convex constraints,
({\em 2}+1)-consistency (path consistency) implies global
consistency of the network.

The usefulness of our new set-based approach is twofold.
Firstly, it gives a clear picture of many of the existing results.
For example, many well known results in the second direction based
on semantic properties of the constraints \cite<including> {vBD95,vBD97},
as well as results from the first direction, can be shown with easy proofs
that make use of set intersection properties.
Secondly, by directing the study of consistency to that of set intersection
properties,
it helps improve some of the existing results
and derive new results as demonstrated in sections 
\ref{sec:set-applicationConvexity}--\ref{sec:set-applicationRC}.


This paper is organized as follows.
In Section \ref{sec:preliminaries}, we present necessary notations 
and concepts.
In Section \ref{sec:set-intersection}, we focus on 
properties of the intersection of {\em tree convex} sets 
and sets with cardinality restrictions.
In Section \ref{sec:set-lift}, we develop a characterization 
of $k$-consistency utilizing set intersection and the
proof schema that offers a generic way to obtain consistency results
from set intersection properties. 
The power of the new approach is demonstrated by new consistency
results on the convexity and tightness of constraints. 
Tree convex constraints are studied 
in Section~\ref{sec:set-applicationConvexity}. 
On a constraint network of tree convex constraints,
local consistency ensures global consistency, as a result of 
the intersection property of tree convex sets. 
The tightness of constraints is studied in
Section~\ref{sec:tightness}.
Thanks to the intersection properties of sets with cardinality
restriction, a relation between local and global consistency 
is identified on {\em weakly tight} constraint networks in 
Section~\ref{sec:consistencyTightness}. These networks
require only some, rather than all, constraints to be $m$-tight, 
improving the tightness result by \citeA{vBD97}. 
With the help of relational consistency, we show that global
consistency can be achieved through local consistency on 
weakly tight constraint networks in 
Section~\ref{sec:globalConsistency}. This type of result on 
tightness was not known before. 
In Section~\ref{sec:propertyWeakTightness}, we explore when a 
constraint network is weakly m-tight and present several  
results about the number of tight constraints sufficient or necessary for 
a network to be weakly tight. To make full use of the tightness of 
the constraints in a network, we 
propose dually adaptive consistency in 
Section~\ref{sec:duallyAdaptive}. Dually adaptive consistency of
a constraint network is determined by its topology 
and the tightest relevant constraint on each variable. 
For completeness, we include in Section~\ref{sec:set-applicationRC} 
results on tightness and tree convexity that are based on
relational consistency.
We conclude in Section~\ref{sec:set-relatedwork}.

\section{Preliminaries }
\label{sec:preliminaries}

A {\em constraint network} \network{R} is defined as a set of 
variables $N = \{x_1, x_2, \ldots, x_n\}$; a set of finite 
domains $D = \{D_1, D_2, \ldots, D_n\}$
where domain $D_i$, for all $i \in 1..n$, is a set of values that 
variable $x_i$ can take; and a set of constraints $C = \{ c_{S_1}, c_{S_2},
\ldots, c_{S_e} \}$ where $S_i$, for all $i \in 1..e$, is a subset of 
$\{x_1, x_2, \ldots, x_n\}$ and each constraint $c_{S_i}$ is a 
relation defined on the domains of all variables in $S_i$. 
Without loss of generality, {\em we assume that, for any two 
constraints $c_{S_i}, c_{S_j} \in C$ ($i \neq j$), $S_i \neq S_j$.}
The {\em arity} of constraint $c_{S_i}$ 
is the number of variables in $S_i$. For a variable $x$, $D_x$ denotes its
domain. In the rest of this paper, we will often use network to mean 
constraint network.

An instantiation of variables $Y = \{x_1, \ldots, x_j\}$
is denoted by $\bar {a} = (a_1, \ldots, a_j)$  where $a_i \in D_i$ for 
$i \in 1..j$. 
An {\em extension} of $\bar{a}$ to a variable $x(\notin Y)$ is denoted by
$(\bar{a},u)$ where $u \in D_x$. 
An instantiation of a set of variables $Y$ is {\em consistent}
if it satisfies all constraints in
\network{R} that do not involve any variables outside $Y$.

A constraint network \network{R} is {\em k-consistent}
if and only if for any consistent
instantiation $\bar {a}$ of any distinct $k-1$ variables, and
for any new variable $x$, there exists
$u \in D_x$ such that $(\bar{a},u)$ is a consistent
instantiation of the $k$ variables.
\network {R} is {\em strongly}
$k$-consistent if and only if it is $j$-consistent for all $j \le k$.
A strongly $n$-consistent network is called {\em globally consistent}.

For more information on constraint networks 
and consistency, the reader is referred to the work by 
Mackworth \citeyear{Mac77}, Freuder \citeyear{Fre78}
and Dechter \citeyear{Dec03}.
\section{Properties on Set Intersection }
\label{sec:set-intersection}

In this section, we develop a number of set intersection results
that will be used later to derive results on consistency.
The set intersection property that we are concerned with is:
\begin{quote}
\it
Given a collection of $l$ finite sets, under what conditions
is the intersection of all $l$ sets not empty?
\end{quote}


Here, we are particularly interested in
the intersection property on sets with two interesting and 
useful restrictions: convexity and cardinality.

\subsection{Tree Convex Sets}

Given a collection of sets, some structures can be associated
with the elements of the sets such that 
we can obtain interesting and useful set intersection results. 
Here we study the sets whose elements form a tree. We first 
introduce the concept of a tree convex set. 

\begin{definition}
Given a discrete set $U$ and a tree \tree\ with vertices $U$,
a set $A \subseteq U$ is {\em \treeconvex} under \tree\ if
there exists a subtree of \tree\ whose vertices 
are $A$.
\end{definition}

A subtree of a tree \tree\ is a subgraph of \tree\ that is 
a tree. Next we define when we can say a collection of sets 
are tree convex. 

\begin{definition}
Given a collection of discrete sets \set, let the union of the sets of 
\set\ be $U$. The sets of \set\ are {\em tree convex under a 
tree \tree\ on $U$} 
if every set of \set\ is tree convex under \tree.
\end{definition}

A collection of sets are said {\em tree convex} if there exists a 
tree such that the sets in the collection are tree convex under
the tree. 

\begin{figure}[ht]
\begin{center}
\input{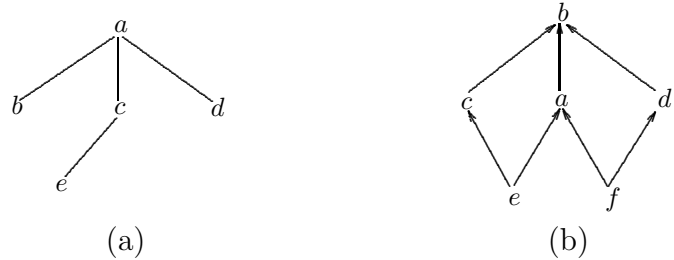}
\end{center}
\caption{(a) A tree with nodes \{$a, b, c, d, e$\} 
(b) A partial order with nodes \{$a, b, c, d, e, f$\} }
\label{fig:set-treeconvex}
\end{figure}

\begin{example} 
Consider a set $U = $ \{$a, b, c, d, e$\} and a tree 
given in \figref{fig:set-treeconvex}. 
The subset \{$a, b, c, d$\} is \treeconvex\ under the given tree.
So is the set \{$b, a, c, e$\} since the elements of the set
are a subtree. However, \{$b, c, e$\} is not \treeconvex\
as its elements do not form a subtree of the given tree.
\end{example}

\begin{example} 
Consider $\set = \{ \{1, 9\}, \{3, 9\}, \{5, 9\} \}$. If we 
construct a tree on $\{ 1, 3, 5, 9\}$ with $9$ being the root
and $1, 3, 5$ being its children,
each set of \set\ covers the nodes of exactly one branch of the tree.
Hence, the sets of \set\ are tree convex.
\end{example}

Tree convex sets have the following intersection property.

\begin{lemma}[Tree Convex Sets Intersection]
  \label{lm:treeConvexSets}
Given a finite collection of finite sets \set, 
assume the sets of \set\ are \treeconvex.
$\bigcap\limits_{E \in \set} E \neq \emptyset$
iff for all $E_1, E_2 \in$ \set,
$E_1 \bigcap\limits E_2 \neq \emptyset$.
\end{lemma}

{\bf Proof.}
Let $l$ be the number of sets in \set, and \tree\ a tree such 
that, for each $E_i \in$ \set, $E_i$ is the vertices of a subtree $T_i$ 
of \tree. Assuming \tree\ is a rooted tree, every $T_i$ ($i \in 1..l$) 
is a rooted tree whose root is exactly the node nearest to the root of \tree.
Let $r_i$ denote the root of $T_i$ for $i \in 1..l$.

To prove $\bigcap\limits_{i \in 1..l} E_i \neq \emptyset$, we want
to show the intersection of the trees $\{ T_i ~|~ i \in 1..l\}$ is
not empty. The following propositions on subtrees are necessary in
our main proof.

\begin{proposition} \label{prop:tree}
Let $T_1, T_2$ be two subtrees of a tree \tree, and $T = T_1 \cap
T_2$. $T$ is a tree.
\end{proposition}
If $T = \emptyset$, it is a trivial tree. Now let $T \neq
\emptyset$. Since $T$ is a portion of $T_1$, there is no circuit
in it. It is only necessary to prove $T$ is connected. That is to
show, for any two nodes $u, v \in T$, there is a path between
them. $u, v \in T_1$ and $u, v \in T_2$ respectively imply that
there exist paths $P_1: u, \ldots, v$ in $T_1$ and $P_2: u,
\ldots, v$ in $T_2$ respectively. Recall that {\em there is a
unique path from $u$ to $v$ in \tree} and that $T_1$ and $T_2$ are
subtrees of \tree. Therefore, $P_1$ and $P_2$ cover the same nodes
and edges, and thus they are in $T$, the intersection of $T_1$ and
$T_2$. $P_1$ is the path we want.

\begin{proposition}  \label{prop:root}
Let $T_1, T_2$ be two subtrees of a tree \tree, and $T = T_1 \cap
T_2$. $T$ is not empty if and only if at least one of the roots of
$T_1$ and $T_2$ is in $T$.
\end{proposition}
Let $r_1$ and $r_2$ be the roots of $T_1$ and $T_2$ respectively.
If $r_1 \in T$, the proposition is correct. Otherwise, we show
$r_2 \in T$. Assume the contrary $r_2 \notin T$. Clearly, $r_1 \neq r_2$.
Let $r$ be the first common ancestor of $r_1$ and $r_2$ and 
$v$ the root of $T$ ($T$ is a tree by
Proposition~\ref{prop:tree}). We have paths $P_1: r_1, \ldots, v$
in $T_1$; $P_2: r_2, \ldots, v$ in $T_2$; and $P_3: r, \ldots,
r_1$, and $P_4: r, \ldots, r_2$ in \tree. 
Since $v$ is a descendant of both $r_1$ and $r_2$, $P_1$ and 
$P_2$ share only the vertex $v$. 
Since $r$ is the first common ancestor of $r_1$ and $r_2$, 
$P_3$ and $P_4$ share only the vertex $r$. It can also be verified 
that $P_3$ and $P_1$ share only $r_1$, $P_2$ and $P_4$ share only
$r_2$, and no vertex is shared by either $P_1$ and $P_4$ or $P_2$ and 
$P_3$. 
Hence, the closed walk $P_3P_1P_2'P_4'$, where
$P_2'$ and $P_4'$ are the reverse of $P_2$ and $P_4$ respectively,
is a simple circuit. 
It contradicts that there is no circuit in \tree.

Further, we have the following observation.
\begin{proposition} \label{prop:maxIsRoot}
Let \tree\ be  a tree with root $r$, and $T_1$ and $T_2$
two subtrees of \tree\ with roots $r_1$ and $r_2$ respectively. 
Let $r_1$ be
not closer to $r$ than $r_2$, and $T$ the intersection of $T_1$
and $T_2$. $r_1$ is the root of $T$ if $T$ is not empty.
\end{proposition}
The proposition is true if $r_1=r_2$. 
Now let $r_1$ be farther to $r$ than $r_2$. 
Clearly $r_2 \notin T_1$ and thus $r_2 \notin T$. 
By Proposition~\ref{prop:root}, $r_1$ is the root of $T$.

\noindent Let $T = \bigcap\limits_{i \in 1..l} T_i$. We are ready
now to prove our main result $ T \neq \emptyset$. Select a tree
$T_{\max}$ from $T_1, T_2, \ldots, T_l$ such that its root
$r_{\max}$ is the farthest away from $r$ of \tree\ among the roots
of the concerned trees. In accordance with
Proposition~\ref{prop:maxIsRoot}, that $T_{\max}$ intersect with
every other tree implies that $r_{\max}$ is a node of every $T_i$
($i \in 1..l$). Therefore, $r_{\max} \in T$. \QED 
\hide{
 \begin{definition}
 A collection of sets \set\ are {\em convex} if
 and only if there exists a total ordering $\preceq$ on
 $D$ such that for every $E_i \in$ \set\
 $E__i = \{ v \in D ~|~ \min E_i \preceq v \preceq \max E_i\}$.
 \end{definition}
}


{\bf Remark.} A partial order can be represented by an
acyclic directed graph. It is tempting to further generalize 
\treeconvexity\ to partial convexity in the following way. \\

{\em Given a set $U$ and a partial order on it, a set $A \subset U$ is
{\em partially convex} if and only if $A$ is the set of nodes of a
connected subgraph of the partial order.
  Given a collection of sets \set, let the union of the
sets of \set\ be $U$. The sets of \set\ are {\em partially convex} if
  there is a partial order on $U$ such that every set of \set\
  is partially convex under the partial order.} \\

However, with this generalization, we can not get a result similar
to Lemma~\ref{lm:treeConvexSets},  which is illustrated
by the following example. Consider three sets
\{$c, b, d$\}, \{$d, f, a$\} and \{$a, e, c$\} that
are the nodes of the diagram given in \figref{fig:set-treeconvex}(b). 
These sets are partially convex and intersect pairwise. 
However, the intersection of all three sets is empty.


\subsection{Sets with Cardinality Restrictions}

Another useful restriction that we will place on sets is to
restrict their cardinalities. As a special case, consider  
a set with only one element $a$. If its intersection with every
other set is not empty, we are able to conclude that every set
contains $a$, and thus the intersection of all the sets is not empty. 
Generally, if a set has at most $m$ elements, we have the
following result. 

\begin{lemma}\label{lm:smallSetM}
Consider a finite collection of $l$ sets 
\set=$\{E_1, E_2, \ldots, E_l\}$ and a number $m < l$. 
Assume one set $E_1 \in$ \set\ has at most $m$ elements.
\[
  \bigcap\limits_{E \in \set} E \neq \emptyset
\]
\noindent iff the intersection of $E_1$ and any other $m$ sets 
of \set\ is not empty.
\end{lemma}

{\bf Proof.}
The necessary condition is immediate.

To prove the sufficient condition, we show that the intersection of  
$E_1$ and any other $k$ ($m \le k \le l-1$) 
sets of \set\ is not empty by induction on $k$.
When $k= m$, the lemma is true according to its assumption.
Assuming that  the intersection of $E_1$ and any other 
$k-1 ~(\ge m)$ sets of \set\ is not empty,
we show that the intersection of $E_1$ and any other $k$ sets of \set\ is not empty. 
Without loss of generality, the
subscripts of the $k$ sets are numbered from $2$ to $k+1$.
For $2 \le i \le k+1$, let $A_i$ be the intersection of $E_1$ and 
the $k$ sets except $E_i$:

\[
  A_i = E_1 \cap \ldots \cap E_{i-1} \cap E_{i+1} \cap \ldots \cap E_{k+1}.
\]

First, we show by contradiction that there exist 
some $i,j \in 2..k+1, i \neq j$ such that $A_i \cap A_j \neq \emptyset$.
Assume $A_i \cap A_j = \emptyset$ for all distinct $i$ and $j$. 
According to the construction of $A_i$'s,
\[E_1 \supseteq \bigcup\limits_{i \in 2..k+1} A_i,\]
and $|A_i| \ge 1$ by the induction assumption.
Hence, 
\[ |E_1| \ge \sum\limits_{i \in 2..k+1} |A_i| \ge k > m, \]
which contradicts $|E_1| \le m$. 

Since $A_i \cap A_j \neq \emptyset$ for some $i,j \in 2..k+1, i \neq j$,
\[
  A_i \cap A_j = \bigcap\limits_{i \in 1..k+1} E_i \neq \emptyset.
\]
\QED

This lemma leads to the following corollary where the intersection
of every $m+1$ sets is not empty. 

\begin{corollary}[Small Set Intersection] \label{lm:smallSet}
Consider a finite collection of $l$ sets \set\ and a number $m < l$. 
Assume one set of \set\ has at most $m$ elements.
\[
  \bigcap\limits_{E \in \set} E \neq \emptyset
\]
\noindent iff the intersection of any $m+1$ sets of \set\ is not
empty.
\end{corollary}

There are other specialized versions \cite{ZY03b} of Lemma~\ref{lm:smallSetM} 
on which some existing works by \citeA{vBD97} and \citeA{Dav93} are based. 

When the sets of concern have a cardinality larger than a certain 
number, the intersection of these sets is not empty under some conditions. 
The reader may refer to the Large Sets Intersection lemma 
\cite{ZY03b} for more details. 

\section{Set Intersection and Consistency}
\label{sec:set-lift}

In this section, we first relate consistency in constraint networks
to set intersection. 
Using this result, we present a proof schema that
allows us to study the relationship between local and global
consistency from the properties of set intersection.

Underlying the concept of $k$-consistency is whether an instantiation
of some variables can be extended to a new variable such that all
relevant constraints on the new variable are satisfied. A {\em
relevant} constraint on a variable $x_i$ {\em with respect to} $Y$
is a constraint that contains only
$x_i$ and some variables of $Y$. 
Given an instantiation of $Y$, each relevant constraint allows 
a set (possibly empty) of values
for the new variable. We call this set an {\em extension set}.
The satisfiability of all relevant constraints depends on
whether the intersection of their extension sets is non-empty (see
Lemma~\ref{lm:setLift}). 

\begin{definition}
Given a constraint $c_{S_i}$, a variable $x \in S_i$, and any
instantiation $\bar {a}$ of $S_i - \{x\}$, the {\em extension set}
of $\bar {a}$ to $x$ with respect to $c_{S_i}$ is defined as
\[
  E_{i,x}(\bar {a}) = \{ b \in D_x ~|~ (\bar {a}, b)
~satisfies~ c_{S_i}\}.
\]
An extension set is {\em trivial} if it is empty; otherwise it is
{\em non-trivial}.
\end{definition}

Recall that $D_x$ refers to the domain of variable $x$.
Throughout the paper, it is often the case that an instantiation
$\bar{a}$ of $S-\{x\}$ is already given, where $S-\{x\}$ is a
superset of $S_i - \{x\}$. Let $\bar{b}$ be the instantiation
obtained by restricting $\bar{a}$ to the variables only in $S_i -
\{x\}$. For ease of presentation, we continue to use $E_{i,x}(\bar
{a})$, rather than $E_{i,x}(\bar {b})$, to denote the extension of
$\bar {b}$ to $x$ under constraint $c_{S_i}$. To make the
presentation easy to follow, some of the three parameters $i$,
$\bar {a}$, and $x$ may be omitted from an expression hereafter
whenever they are clear from the context. For example, given an
instantiation $\bar{a}$ and a new variable $x$, to emphasize
different extension sets with respect to different constraints
$c_{S_i}$, we write $E_i$ instead of $E_{i,x}(\bar{a})$ to simply
denote an extension set.

\begin{example} 
\label{eg:extension-set}
Consider a network with variables \{$x_1, x_2, x_3, x_4, x_5$\}:
\[
\begin{array}{c}
\begin{array}{rcll}
 c_{ S_1} &=& \{ (a, b, d), (a, b, a) \},&S_1 = \{x_1, x_2, x_3 \};\\
 c_{ S_2} &=& \{ (b, a, d), (b, a, b) \},&S_2 = \{x_2, x_4, x_3 \};\\
 c_{ S_3} &=&  \{(b,d), (b, c) \}, &S_3 = \{x_2, x_3 \};\\
 c_{ S_4} &=& \{ (b, a, d), (b, a, a) \}, &S_4 = \{x_2, x_5, x_3 \};\\
\end{array} \\
 D_1 = D_4 = D_5 = \{a\}, D_2 = \{b\}, D_3 = \{ a, b, c, d\}.
\end{array}
\]
Let $\bar{a} = (a,b,a)$ be an instantiation of variables
$Y = \{x_1, x_2, x_4\}$.
The relevant constraints to $x_3$ are $c_{S_1}$, $c_{S_2}$, and $c_{S_3}$.
$c_{S_4}$ is not relevant since it contains $x_5$ outside $Y$.
The extension sets of $\bar{a}$ to $x_3$ with respect to the relevant
constraints are:
\[
  E_1(\bar{a}) = \{d, a\},  E_2(\bar{a}) = \{d, b\}, E_3(\bar{a}) = \{d, c\}.
\]
The intersection of the extension sets above is not empty,
implying that $\bar{a}$ can
be extended to satisfy all relevant constraints $c_{S_1}, c_{S_2}$
and $c_{S_3}$.

\hide{
 \footnote{Of course, the definition of extension set can be generalized to
 accommodate this case. See the section of application[---??locate the name for
 a section where this generalization is discussed---]}
}
Let $\bar{a} = (b, c)$ be an
instantiation of $\{x_2, x_3\}$. $E_{1,x_1}(\bar{a}) = \emptyset$ and
thus it is trivial. In other words, with a trivial
extension set, an instantiation can not be extended to satisfy 
the constraint of concern.
\end{example}

The relationship between {\em k-consistency} and set intersection
is characterized by the following lemma.

\begin{lemma}[Set Intersection and Consistency; Lifting] \label{lm:setLift}
A constraint network \network{R} is $k$-consistent if and only if for any
consistent instantiation $\bar{a}$ of any $(k-1)$ distinct variables
$Y = \{ x_1, x_2, \ldots, x_{k-1} \}$, and any new variable $x_k$,
\[ \bigcap\limits_{j \in 1..l}E_{i_j} \neq \emptyset \]
\noindent where $E_{i_j}$ is the extension set of $\bar {a}$ to
$x_k$ with respect to $c_{S_{i_j}}$, and $c_{S_{i_1}}, \ldots,
c_{S_{i_l}}$ are all relevant constraints.
\end{lemma}

{\bf Proof.} It follows directly from the definition of
$k$-consistency in Section~\ref{sec:preliminaries} and the
definition of extension set. \QED

The insight behind this lemma is to examine consistency from the
perspective of set intersection.

\begin{example} Consider again Example \ref{eg:extension-set}. 
We would like
to check whether the network is $4$-consistent. Consider the
instantiation $\bar{a}$ of $Y$ again. This is a trivial consistent
instantiation since the network doesn't have a constraint among
the variables in $Y$. To extend it to $x$, we need to check the
first three constraints $c_{S_1}$ to $c_{S_3}$. 
The extension is feasible because the
intersection of $E_1, E_2$, and $E_3$ is not empty. We show the
network is $4$-consistent, by exhausting all consistent
instantiations of any three variables. Conversely, if we know the
network is $4$-consistent, we can immediately say that the
intersection of the three extension sets of $\bar{a}$ to $x$ is
not empty. 
\end{example}

The usefulness of this lemma is that it allows consistency
information to be obtained from the intersection of extension
sets, and vice versa. With this point of view of consistency as set
intersection, some results on set intersection properties,
including all those in Section~\ref{sec:set-intersection}, can be
{\em lifted} to get various consistency results for a constraint
network through the following {\em proof schema}.

{\bf Proof Schema}

1. ({\em Consistency to Set}) From a certain level of consistency
{\em in} the constraint network,
we derive information on the intersection of the extension sets
by Lemma~\ref{lm:setLift}.

2. ({\em Set to Set}) From the {\em local} intersection information of sets,
information may be obtained on intersection of more sets.

3. ({\em Set to Consistency}) From the new information on intersection of
extension sets, higher level of consistency is obtained by
Lemma~\ref{lm:setLift}.

4. ({\em Formulate conclusion on the consistency of the constraint network}).
\QED

\noindent In the proof schema, step 1 (consistency to set), step 3 
(set to consistency), and step 4 are straightforward in many cases. 
So, Lemma~\ref{lm:setLift} is also called the {\em lifting} lemma because
once we have a set intersection result (step 3), we can easily have 
consistency results on a network (step 4). The proof schema establishes
a direct relationship between set intersection and consistency properties
in a constraint network.

In the following sections, we demonstrate how the set intersection
properties and the proof schema are used to obtain new 
results on the consistency of a constraint network.

\section{Global Consistency of Tree Convex Constraints}
\label{sec:set-applicationConvexity}

The notion of {\em extension set} plays the role of a bridge
between the restrictions to set(s) and properties of special constraints. 
In this section,
we consider the constraints arising from tree convex sets 
(Lemma~\ref{lm:treeConvexSets}).
A constraint is {\em tree convex} if
all the extension sets with respect to the constraint are tree convex.

\begin{definition}
A constraint $c_S$ is {\em \treeconvex} with respect to $x_i$ and 
a tree $T_i$ on $D_i$ if and only if the sets in
\begin{center}
$A=\{ E_{S,x_i} ~|~ E_{S,x_i}$ is a non-trivial extension of some
instantiation of $S-\{x_i\} \}$
\end{center}
are \treeconvex\ under $T_i$.
A constraint $c_S$ is {\em tree convex} under a tree $T$
on the union of the domains of the variables in $S$, if it is 
tree convex with respect to every $x \in S$ under $T$.
\end{definition}

\hide{
\begin{figure}[ht]
\begin{center}
\begin{tabular}{cc}
\begin{minipage}{2.0in}
\small
\begin{center}
\begin{tabular}{c|cccc}
  & 1 & 3 & 5 & 9 \\ \hline
1 & * &   &   & * \\
3 &   & * &   & * \\
5 &   &   & * & * \\
9 & * & * & * & * \\
\end{tabular}
\end{center}
\end{minipage}
&
\begin{minipage}{2.0in}
\begin{center}
\includegraphics{tree-constraint.eps}
\end{center}
\end{minipage} \\
(a) & (b)
\end{tabular}
\end{center}
\caption{A tree convex constraint (a) and a tree (b) on its domain}
\label{fig:tree-constraint}
\end{figure}
}

\begin{example} \label{ex:accessibility} 
Tree convex constraints can occur where there
is a relationship among the values of a variable. Consider
the constraint on the accessibility of a set of facilities 
by a set of persons. The personnel include a network engineer,
web server engineer, application engineer, and a team leader.
The relationship among the staff is that the team leader manages
the rest, which forms a tree structure shown in 
\figref{fig:system}(b).
There are  different accessibilities to a system which includes 
basic access, access to the network routers, access to the
web server, and access to the file server. In order to access 
the routers and servers, one has to have the basic access right,
implying a tree structure (\figref{fig:system}(c))
on the access rights. The constraint is that the team leader is
able to access all the facilities while each engineer can
access only the corresponding facility (e.g., the web server engineer
can access the web server). This tree convex constraint
is shown in \figref{fig:system}(a) where the rows are named by
the initials of the engineers and the columns by the initials of
the access rights. The tree on the union of personnel and the
accessibilities can be obtained from their respective trees 
(in \figref{fig:system}(b) and (c)) by adding an edge, 
say between web server and leader. Note that the constraint 
in \figref{fig:system}(a) is not row convex. 

\begin{figure}[ht]
\begin{center}
\begin{tabular}{ccc}
\begin{minipage}{1.2in}
\small
\begin{center}
\begin{tabular}{c|cccc}
  & r & w & f & b \\ \hline
n & * &   &   & * \\
w &   & * &   & * \\
a &   &   & * & * \\
l & * & * & * & * \\
\end{tabular}
\end{center}
\end{minipage}
&
\begin{minipage}{1.5in}
\begin{center}
\includegraphics{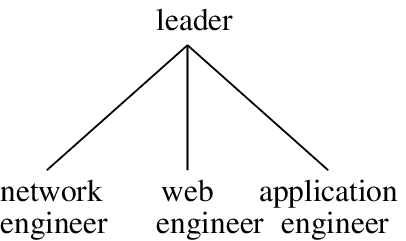}
\end{center}
\end{minipage}
&
\begin{minipage}{1.5in}
\begin{center}
\includegraphics{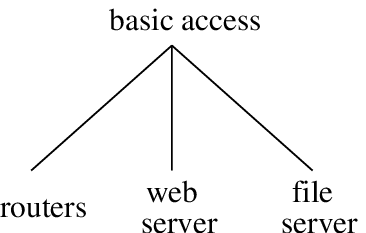}
\end{center}
\end{minipage} \\
(a) & (b) & (c)
\end{tabular}
\end{center}
\caption{A tree convex constraint between accessibilities and staffs}
\label{fig:system}
\end{figure}
\end{example}

\begin{example} 
Tree convex constraints can also be used to model some scene 
labeling problems naturally as shown by \citeA{ZF04a}.  
\end{example}

\begin{definition}
A constraint network is {\em \treeconvex} if 
there exists a tree \tree\ on the union of all its variable 
domains such that all constraints are tree convex under \tree.
\end{definition}

Tree convex constraints generalize row convex constraints 
introduced by \citeA{vBD95}.

\begin{definition}
A constraint $c_S$ is {\em row convex} with respect to $x$
if and only if the sets in
\begin{center}
$A=\{ E_{S,x} ~|~ E_{S,x}$ is a non-trivial extension of some instantiation of $S-\{x\} \}$
\end{center}
are tree convex under a tree where any node has at most one child.
Such a tree is called a {\em total ordering}.
A constraint $c_S$ is row convex if, 
under a total ordering on the union of the involved
domains, it is row convex with respect to every $x \in S$.
\end{definition}

\begin{example} For the constraint $c$ 
in Example~\ref{ex:accessibility} to be row convex,
b (basic access) has to be the neighbor of r (routers), 
w (web server), and f (file server). However, in a total
ordering, a value can be the neighbor of at most two other
values. Hence, $c$ is not row convex but is tree convex.
\end{example}

By the property of set intersection on tree convex 
sets and the proof schema,
we have the following consistency results on tree convex
constraints. 

\begin{theorem}[Tree Convexity] \label{th:treeConvexity}
Let \network{R} be a network of constraints 
with arity at most $r$ and strongly $2(r-1)+1$ consistent.
If \network{R} is tree convex then it is globally consistent.
\end{theorem}

{\bf Proof.} The network is strongly $2(r-1)+1$ consistent by assumption.
We prove that the network is $k$ consistent for any
$k \in \{2r, \ldots, n\}$.

Consider any instantiation $\bar{a}$ of any $k-1$ variables and any
new variable $x$. Let the number of relevant constraints be $l$.
For each relevant constraint, there is one extension set of $\bar{a}$ to $x$.
So, we have $l$ extension sets. If the intersection of all $l$ sets is
not empty, we have a value for $x$ such that the extended instantiation
satisfies all relevant constraints.

({\em Consistency to Set})
Consider any two of the $l$ extension sets: $E_1$ and $E_2$.
The two corresponding constraints involve at most $2(r-1)+1$ variables
since the arity of a constraint is at most $r$ and each of the two constraints
has $x$ as a variable.
By the consistency lemma, that \network {R} is
$(2(r-1)+1)$-consistent implies that the intersection of $E_1$ and $E_2$ is
not empty.

({\em Set to Set})
Since all relevant constraints are \treeconvex\ under the given tree,
the extension sets of $\bar{a}$ to $x$ are tree convex.
Henceforth, the fact that every two of the extension sets intersect
shows that the intersection of all $l$ extension sets is not
empty, by the {\em \treeconvex\ sets intersection} lemma.

({\em Set to Consistency})
From the consistency lemma, we have that \network {R} is $k$-consistent.
\QED

Since a row convex constraint is tree convex, this result generalizes the
consistency result on row convex constraints reported by
\citeA{vBD95}. It is interesting to observe that the latter can be 
lifted from a set intersection results on convex sets \cite{ZY03b}. 

A question raised by Theorem~\ref{th:treeConvexity} 
is how efficient it is to check whether a constraint 
network is tree convex. \citeA{Yos03} has proposed 
an algorithm to recognize a tree convex constraint network 
in polynomial time.

\section{Consistency and the Tightness of Constraints}
\label{sec:tightness}

In this section, we will present various consistency results 
on the networks with $m$-tight constraints. 

\subsection{Global Consistency on Weakly Tight Networks}
\label{sec:consistencyTightness}

The tightness of constraints has been related to 
the consistency of a constraint network by \citeA{vBD97}. 
The {\em m-tightness} of a constraint is characterized by
the cardinality of the extension sets in the following way.

\begin{definition} {\em \cite{vBD97}}
A constraint $c_{S_i}$ is {\em m-tight} with respect to $x \in S_i$
iff for any instantiation $\bar {a}$ of $S_i - \{x\}$,
\[ |E_{i,x}| \le m ~ or ~ |E_{i,x}| = |D_x|. \]
\noindent A constraint $c_{S_i}$ is {\em m-tight} iff it is
m-tight with respect to every $x \in S_i$.
\end{definition}

Given an instantiation, if its extension set with respect to
$x$ is the same as the domain of variable $x$, i.e., $|E_{i,x}| = |D_x|$, 
the instantiation is supported by all values of $x$ and thus 
easy to be satisfiable. Hence, in the definition above, these 
instantiations do not affect the $m$-tightness of a constraint.  

\begin{figure}[htb!]
\begin{center}
\input{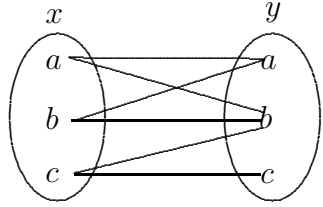}
\end{center}
\caption{The constraint $c_{xy}$ is  2-tight or 3-tight}
\label{fg:2tight}
\end{figure}

\begin{example}
Consider the constraint $c_{xy}$ in \figref{fg:2tight} where 
$D_x=D_y=\{a,b,c\}$. An edge in the graph denotes that
its ends are allowed by $c_{xy}$. It can be verified that 
for the values of $x$, their extension sets have a cardinality of $2$,
and for values of $y$, their extension sets have a cardinality from $1$ to $3$. 
Hence, $c_{xy}$ can be said 2-tight or 3-tight but not 1-tight.
\end{example}

We are specially interested in the following tightness.

\begin{definition}
A constraint $c_{S_i}$ is {\em properly m-tight} with respect to $x \in S_i$
iff for any instantiation $\bar {a}$ of $S_i - \{x\}$,
\[ |E_{i,x}| \le m. \]
\noindent A constraint $c_{S_i}$ is {\em properly m-tight} iff it is
properly m-tight with respect to every $x \in S_i$.
\end{definition}

A constraint is $m$-tight if it is properly $m$-tight. The
converse might not be true. For example, the constraint $x \le y$,
where $x \in \{1, 2, \ldots, 10\}$ and $y \in \{1, 2, \ldots, 10
\}$, is $9$-tight but not properly $9$-tight. It is properly
$10$-tight since $|E_{x}(10)| = 10$ when $y=10$.

Next, we define a special constraint network which allows us to 
make a more accurate connection between the tightness of 
constraints and the consistency of the network. 

\begin{definition} \label{def:weaklyM-tight} A constraint network 
is {\em weakly} m-tight at level $k$ iff for every set of 
variables $\{x_1, x_2, \ldots, x_l\} (k \le l < n) $ and a new variable $x$,
there exists a properly
$m$-tight constraint among the relevant constraints on $x$ with
respect to $\{x_1, x_2, \ldots, x_l\}$.
\end{definition}

\begin{figure}[htb!]
\begin{center}
\input{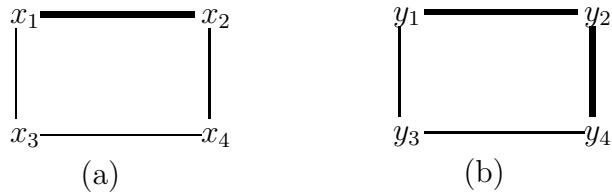}
\end{center}
\caption{Two constraint networks. A thin edge represents a properly
$m$-tight constraint while a thick one represents a non properly 
$m$-tight constraint}
\label{fg:weaklyTightNetwork}
\end{figure}

\begin{example} \label{eg:weaklyTightNetwork}
The network in \figref{fg:weaklyTightNetwork}(a) is weakly tight at level 
3 because for any three variables and a fourth variable, one of the 
relevant constraints is properly $m$-tight. The network in 
\figref{fg:weaklyTightNetwork}(b) is not weakly tight at level 3 since 
for \{$y_1$, $y_3$, $y_4$\} and $y_2$, none of the relevant constraints
 $c_{y_1y_2}$ and $c_{y_4y_2}$ is properly $m$-tight.  
\end{example}

By the small set intersection corollary (Corollary \ref{lm:smallSet}),
we have the following consistency result on a weakly $m$-tight network.

\begin{theorem}[Weak Tightness] \label{th:weakTightness}
  If a constraint network \network{R}  with
  constraints of arity at most $r$ is strongly $((m+1)(r-1)+1)$-consistent
  and weakly $m$-tight at level $((m+1)(r-1)+1)$, it is globally consistent.
\end{theorem}
{\bf Proof.} Let $j = (m+1)(r-1)+1$. The constraint network
\network {R} will be shown to be $k$-consistent for all $k$ ($j <
k \le n$).

Let  $Y = \{x_1, \ldots, x_{k-1} \}$ be a set of any $k-1$
variables, and $\bar {a}$ be an instantiation of all variables in
$Y$. Consider any additional variable $x_k$. Without loss of
generality, let the relevant constraints be $c_{S_1}, \ldots,
c_{S_l}$, and $E_i$ be the extension set of $\bar {a}$ to $x_k$
with respect to $c_{S_i}$ for $i \le l$.
\smallskip

({\em Consistency to Set}) Consider any $m+1$ of the $l$ extension
sets. All the corresponding $m+1$ constraints contain at most
$(m+1)(r-1)+1$ variables including $x_k$. Since \network{R} is
$((m+1)(r-1)+1)$-consistent, by the {\em set intersection
and consistency} lemma, the intersection of the $m+1$ extension sets
is not empty.
\smallskip

({\em Set to Set}) The network is weakly $m$-tight at level
$((m+1)(r-1)+1)$. So, there must be a properly $m$-tight constraint among
the relevant constraints $c_{S_1}, \ldots, c_{S_l}$. Let it be
$c_{S_i}$. We know its extension set $|E_i| \le m$. 
Since the intersection of every $m+1$ of the extension sets is not empty,
all $l$ extension sets share a common element by the {\em
small set intersection} corollary.
\smallskip

({\em Set to Consistency}) By the lifting lemma, \network {R} is
$k$-consistent. \QED

In a similar fashion, the main tightness result by \citeA{vBD97},
where all the constraints are required to be $m$-tight,
can be lifted from the small sets intersection corollary by 
\citeA{ZY03b}. This uniform treatment of lifting set intersection
results to consistency results is absent from
the existing works \cite<e.g.,>{Dec92b,vBD95,vBD97,Dav93}.

The tightness result by \citeA{vBD97} requires every 
constraint to be $m$-tight.
The weak tightness theorem, on the other hand, does not require
all constraints to be properly $m$-tight.
The following example illustrates this difference.

\begin{table}[htb]
  \begin{center}
    \begin{tabular}{|c||llllllllll|} \hline
     Extension  & \multicolumn{10}{|c|}{Relevant constraints} \\ \hline
     $1234 \ra 5$, & 125*,& 135~ ,  & 145~ ,& 235,& 245, & 345,& 15+,& 25~ ,& 35~ ,& 45 \\ \hline
     $2345 \ra 1$, & 231~ ,& 241~ , & 251*,& 341,& 351, & 451,& 21~ ,& 31~ ,& 41~ ,& 51+ \\ \hline
     $3451 \ra 2$, & 132~ ,& 142~ , & 152*,& 342,& 352, & 452,& 12~ ,& 32+,& 42~ ,& 52 \\ \hline
     $4512 \ra 3$, & 123~ ,& 143*,& 153~ ,& 243,& 253, & 453,& 13~ ,& 23+,& 43~ ,& 53 \\ \hline
     $5123 \ra 4$, & 124~ ,& 134*,& 154~ ,& 234,& 254, & 354,& 14~ ,& 24~ ,& 34+,& 54 \\ \hline
   \end{tabular}
 \end{center}
 \caption{ \label{wt:tightConstraints} Relevant constraints in extending an
  instantiation of four variables to a new variable}
\end{table}

\begin{example} For a weakly $m$-tight network, we are interested
in its topological structure. Thus we have omitted the domains of
variables here. Consider a network with five variables labeled
$\{1, 2, 3, 4, 5\}$. In this network, for any pair of variables
and for any three variables, there is a constraint.
Assume the network is already strongly $4$-consistent.

Since the network is already strongly $4$-consistent, we can simply
ignore the instantiations with less than $4$ variables. This
is why we introduce the level at which the network is weakly $m$-tight.
The interesting level here is $4$.
Table \ref{wt:tightConstraints} shows the relevant constraints
for each possible extension of four instantiated variables to the other one.
In the first row,
$1234 \rightarrow 5$ stands for extending the instantiation of variables
$\{1,2,3,4\}$ to variable $5$.
Entries in its second column denote a constraint.
For example, 125 denotes $c_{125}$.
If the constraints on
$\{1, 2, 5\}$ and $\{1, 3, 4\}$ (suffixed by * in the table)
are properly $m$-tight, the network is weakly $m$-tight at level $4$.
Alternatively, if the constraints $\{1, 5\}$, $\{2, 3\}$ and $\{3, 4\}$
(suffixed by +) are properly $m$-tight, the network will also be 
weakly $m$-tight. The tightness result by \citeA{vBD97} requires
all binary and ternary constraints to be $m$-tight.
\end{example}

\subsection{Making Weakly Tight Networks Globally Consistent}
\label{sec:globalConsistency}

Consider the weak tightness theorem in
the previous section. Generally, a weakly
$m$-tight network might not have the level of local consistency
required by the theorem. It is tempting to enforce such a level of
consistency on the network to make it globally consistent.
However, this procedure may result in constraints with higher
arity.

\begin{example}
Consider a network with variables $\{x, x_1, x_2,
x_3\}$. Let the domains of $x_1, x_2, x_3$ be $\{ 1, 2, 3\}$, the
domain of  $x$ be $\{1, 2, 3, 4 \}$, and the constraints be that
all the variables should take different values:
$x \neq x_1, x \neq x_2, x \neq x_3, x_1 \neq x_2, x_1 \neq x_3,
   x_2 \neq x_3.
$
This network is strongly path consistent. In checking the
4-consistency of the network, we know that the instantiation
$(1,2,3)$ of $\{x_1, x_2, x\}$ is consistent but can not be
extended to $x_3$. To enforce 4-consistency, it is necessary to
introduce a ternary constraint on $\{x_1, x_2, x\}$ to make $(1,2,3)$ no
longer a valid instantiation.

To make the new network globally consistent, the newly introduced
constraints with higher arity may in turn require higher local
consistency in accordance with Theorem~\ref{th:weakTightness}.
Therefore, it is difficult to predict an exact level of consistency
(variable based) to {\em enforce} on the network to make it
globally consistent.
\end{example}

In this section, relational consistency will be used to make a
constraint network globally consistent.

\begin{definition} {\em \cite{vBD97}}
A constraint network \network\ is {\em relationally m-consistent}
iff given
(1) any m distinct constraints $c_{S_1}, \ldots, c_{S_m}$, and
(2) any $x \in \cap_{i=1}^{m}S_i$, and
(3) any consistent instantiation
$\bar{a}$ of the variables in $(\cup_{i=1}^{m}S_i - \{ x \})$,
there exists an extension of $\bar{a}$ to $x$ such that the extension
is consistent with the $m$ relations. A network is {\em strongly}
relationally $m$-consistent if it is relationally $j$-consistent
for every $j \le m$.
\end{definition}

Variables are no longer of concern in relational consistency.
Instead, constraints are the basic unit of consideration.
Intuitively, relational $m$-consistency concerns whether all $m$
constraints agree at every one of their shared variables. It makes
sense because different constraints interact with each other
exactly through the shared variables.

Relational $1$-, and $2$-consistency are also called
relational arc, and path consistency, respectively.

Using relational consistency, we are able to obtain global
consistency by enforcing local consistency on the network. 

\begin{proposition} \label{prop:preserveTightness}
The weak $m$-tightness at level $k$ of a constraint network is preserved 
by the process of enforcing relational consistency on the network.
\end{proposition}
{\bf Proof.}
Let \network{R} be the constraint network before relational 
consistency enforcing 
and \network{R}$_1$ the network after consistency enforcing.
Clearly, \network{R} and \network{R}$_1$ have the same set of variables. 
Consider any set of variables \{$x_1, x_2, \ldots, x_l$\} ($k \le l < n$)
and a new variable $x$.
Since \network{R} is weakly $m$-tight at level $k$, there exists a 
properly $m$-tight constraint $c$ among the relevant constraints 
on $x$ with respect to \{$x_1, x_2, \ldots, x_l$\}. 
Enforcing relational consistency on a constraint network
will only tighten a constraint. So, the proper $m$-tightness of 
$c$ is preserved. 
Hence, \network{R}$_1$ is weakly $m$-tight at level $k$. 
\QED

Now we have the main result of this subsection.

\begin{theorem} \label{th:binaryMTightNetwork}
A constraint network weakly $m$-tight at level $(m+1)(r-1)+1$,
where $r$ is the maximal arity of the constraints of the network, 
is globally consistent after 
it is made strongly relationally $(m+1)$-consistent.
\end{theorem}

{\bf Proof.} By Proposition~\ref{prop:preserveTightness}, 
the network is still weakly $m$-tight at $(m+1)(r-1)+1$ 
after enforcing strong relational $(m+1)$-consistency on it. 
Let $r_1$ be the maximal arity of the constraints of the 
new network after consistency enforcing. Clearly, 
$r_1 \ge r$. So, the network is $m$-tight at $(m+1)(r_1-1)+1$
by Proposition~\ref{prop:weaknessLevel}. 
The theorem follows immediately from Theorem~\ref{th:weakTightnessRC} in 
Section~\ref{sec:set-applicationRC}. \QED

The implication of this theorem is that as long as we have certain
properly $m$-tight constraints on certain combinations of
variables, the network can be made globally consistent by
enforcing relational $(m+1)$-consistency. 

We have the following observation on the weak $m$-tightness
of a network.

\begin{proposition} \label{prop:weaktightness}
A constraint network is weakly $m$-tight at any level 
if the constraint between
every {\em two} variables in the network is properly $m$-tight.
\end{proposition}

{\bf Proof.} Consider any level $k$, any set of variables $Y = \{
x_1, x_2, \ldots, x_l\} (k \leq l \leq n)$, and any new variable
$x \notin Y$. Since the constraint between any two variables is
properly $m$-tight, the constraint $c_{\{x_1,x\}}$ on $x_1$ and
$x$ is properly $m$-tight. Therefore, there is a properly m-tight
constraint $c_{\{x_1,x\}}$ among the relevant constraints after an
instantiation of $Y$. \QED

This observation shows that the proper $m$-tightness of 
the constraints on every two variables is sufficient 
to determine the level of local consistency
needed to ensure global consistency of a constraint network. 

{\bf Remark.} Proposition~\ref{prop:weaktightness}
assumes there is a 
constraint between every two variables. If there is no
constraint between some two variables, a universal 
constraint is introduced. In this case, we can enforce 
path consistency on the constraint network 
to make the binary constraints tighter so that lower level 
of relational consistency is needed to make the network
globally consistent. 

\subsection{Properties of Weakly Tight Constraint Networks}
\label{sec:propertyWeakTightness}

Since for a weakly $m$-tight constraint network global consistency
can be achieved through local consistency, it is interesting and important
to investigate the conditions for a network to be weakly $m$-tight. 
Although Proposition~\ref{prop:weaktightness} shows a sufficient
condition, it requires every binary constraint be 
tight. As we can see from Example~\ref{eg:weaklyTightNetwork}(a), 
the required number of tight constraints for a constraint network 
to be weakly tight can be further reduced. 
This subsection is focused on the understanding of the relationship
between the number of tight constraints and the weak tightness of a 
constraint network. 

There is a strong relationship among different levels of weak tightness 
in a network.

\begin{proposition} \label{prop:weaknessLevel}
If a constraint network is weakly $m$-tight at level $k$ for some $m$, 
it is weakly $m$-tight at any level $j > k$. 
\end{proposition}

{\bf Proof.} For any $j > k$, we prove that the network 
is weakly tight at level $j$. That is, 
for any set of variables $Y=\{x_1, \ldots, x_j\}
(k \le j < n) $ and a new variable $x$,
we show that there exists an $m$-tight relevant constraint 
 on $x$ with respect to $Y$. Since the network is weakly tight
for $k < j$, there exists an $m$-tight relevant constraint 
on $x$ with respect to a subset of $Y$. This constraint is still relevant
on $x$ with respect to $Y$, and thus the one we look for. 
\QED

In the following, we present two results on sufficient conditions for 
a constraint network to be weakly $m$-tight. 

\begin{theorem} \label{th:ensureTightness}
Given a constraint network $(V, D, C)$ and a number $m$, 
if for every $x \in V$, there are at least $n-2$ properly $m$-tight 
binary constraints on it, then the network is weakly $m$-tight 
at level $2$. 
\end{theorem}

{\bf Proof.} 
For any two variables $\{x, y\}$ and a third variable $z$, 
the relevant constraints on $z$ with respect to $\{x,y\}$ are 
$c_{xz}$ and $c_{yz}$.  
We know that the number of relevant binary constraints on 
$z$ with respect to $V$ is $n-1$. That $n-2$ of them are properly $m$-tight 
means either $c_{xz}$ or $c_{yz}$ must be properly $m$-tight. 
\QED

In fact, for the weak tightness at a higher level, we need fewer constraints
to be $m$-tight as shown by the following result.

\begin{theorem}
A constraint network $(V, D, C)$ is weakly $m$-tight at level $k$
if for every $x \in V$, 
there are at least $n-k$ properly $m$-tight binary constraints 
on it.
\end{theorem}

{\bf Proof.}
For any set $Y$ of $k$ variables and a new variable $z$,
we show that there is a properly $m$-tight relevant constraint on $z$ 
with respect to $Y$. 
Otherwise, none of the $k$ binary constraints on $z$ is 
properly $m$-tight. Since the total number 
of the relevant binary constraints on $z$ is $n-1$,  
the number of properly $m$-tight binary constraints on 
$z$ is at most $(n-1) - k$, which contradicts 
that $z$ is involved in $n-k$ properly $m$-tight binary constraints. 
\QED

This result reveals that for a constraint network to
be weakly tight at level $k$, it could need as few as $n(n-k+1)/2$
properly $m$-tight binary constraints, in contrast to the result in 
Theorem~\ref{th:binaryMTightNetwork}
where all binary constraints are required to be properly $m$-tight.

An immediate question is:
What is the minimum number of $m$-tight constraints required 
for a network to be weakly tight? 
It can be answered by the following result on weak tightness 
at level $2$.



\begin{theorem}
Given a number $m$, for a constraint network to be weakly $m$-tight at 
level $2$, it needs at least 
\[ n(n-1)/2 - 2\lfloor n/3 \rfloor ~~ \mbox{if} ~~
   n = 0,1 ~~\mbox{({\tt mod}} ~ 3 \mbox {)} 
\] or otherwise
\[
   (n-2)(3n-1)/6
\]
$m$-tight binary or ternary constraints.
\end{theorem}

{\bf Proof.}
\newcommand{\tc}[1] {$\{#1\}$}
\newcommand{\bc}[1] {$\{#1\}$}
\newcommand{\networkn}[1] {\network{R}$_#1$}
Given a network, its weak $m$-tightness at level $2$ 
depends on the tightness of only binary and ternary constraints. 
Among all weakly $m$-tight (at level 3) constraint networks with $n$ 
variables, let \networkn{1} be the network that has a minimal set of 
properly $m$-tight binary and ternary constraints.

In the following exposition, a constraint is denoted by its scope.
For example, we use \tc{u,v,w} and \bc{u,v} to
denote ternary constraint $c_{\{u,v,w\}}$ and binary constraint
$c_{uv}$ respectively. A constraint is {\em non-properly-m-tight}
if it is not properly $m$-tight.  

The proof consists of three steps. 

Step 1. While preserving the weak $m$-tightness of \networkn{1} and 
the number of properly $m$-tight constraints in \networkn{1}, 
we modify, if necessary, the proper $m$-tightness of some
constraints in \networkn{1} such that,
for any properly weak $m$-tight constraint \tc{u,v,w},
none of the binary constraints \bc{u,v}, \bc{v,w},
and \bc{u,w} is properly $m$-tight.

To {\em modify} the proper
$m$-tightness of a constraint $c$ in \networkn{1} is to
remove $c$ from the network and introduce a new
constraint on the same set of variables of $c$ with the
desirable proper $m$-tightness. 

We claim that, for any properly $m$-tight constraint
\tc{u,v,w}, at most one of \bc{u,v}, \bc{v,w}, and \bc{u,w} 
is properly $m$-tight.
Otherwise, at least two of them are properly $m$-tight, which means 
\tc{u,v,w} can be modified to be not properly $m$-tight, 
contradicting the minimality of the number of properly $m$-tight
constraints in \networkn{1}.

\begin{figure}[htb!]
\begin{center}
\input{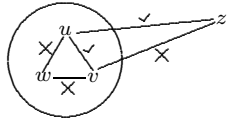}
\end{center}
\caption{\label{fig:nonTight} The circle represents the properly $m$-tight 
ternary constraint \tc{u,v,w}. An edge between two variables indicates a 
binary constraint. A tick besides an edge means it is properly $m$-tight 
while a cross means it is not. 
}
\end{figure}

Assume \bc{u,v} is properly $m$-tight.
Since \tc{u,v,w} is properly $m$-tight, there should be a reason 
for \bc{u,v} to be properly $m$-tight. The only reason is that
there exists another variable $z$ such that one of \bc{u, z} and 
\bc{v, z} is not properly $m$-tight, and \tc{u,v,z} is not properly $m$-tight, 
too. See \figref{fig:nonTight}.
Without loss of generality, let \bc{u,z} be properly $m$-tight,
implying that the constraint \bc{v,z} is not properly $m$-tight. 
The constraint \tc{z, v, w} is properly $m$-tight because
\bc{v,z} and \bc{v,w} are not properly $m$-tight. 

Now we modify the constraints \tc{u,v,w} and \tc{z,v,w} to
be not properly $m$-tight and modify the constraints 
\bc{z,v} and \bc{v,w} to be properly $m$-tight.
This modification preserves the number of properly $m$-tight constraints
in \networkn{1} and the weak $m$-tightness of \networkn{1}.

Step 2. While preserving the weak $m$-tightness of \networkn{1} and 
the number of properly $m$-tight constraints in \networkn{1}, 
we next modify, if necessary, the proper $m$-tightness of the 
constraints in \networkn{1} such that
no two properly $m$-tight ternary constraints 
share any variables.

\begin{figure}[htb!]
\begin{center}
\includegraphics{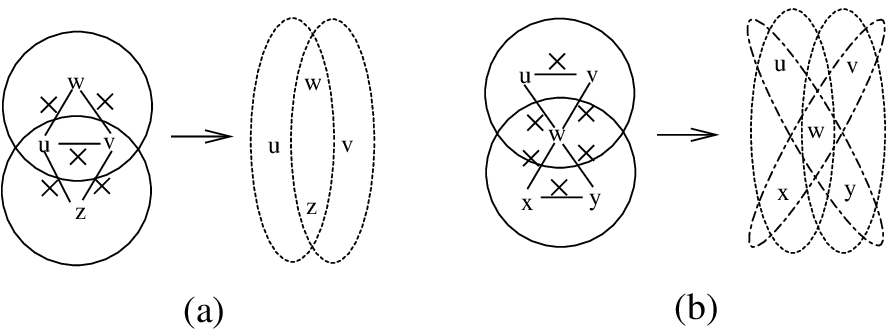}
\end{center}
\caption{\label{fig:2TernaryC} A dotted ellipse together with the three
variables inside it represents a ternary constraint.
(a) Left: Two ternary constraints share two variables \bc{u,v}. Right:
The ternary constraints have to be properly $m$-tight.
(b) Left: Two ternary constraints share one variable $w$.
Right: The ternary constraints have to be properly $m$-tight.
}
\end{figure}

Case 1: Two properly $m$-tight constraints \tc{u,v,w} and \tc{u,v,z} 
share two variables \{$u,v$\}. 
See \figref{fig:2TernaryC}(a).
Since \bc{w,u} and \bc{u,z} are not properly $m$-tight (in terms of 
step 1), \tc{w,u,z} has to be properly $m$-tight.
Since \bc{w,v} and \bc{v,z} are not $m$-tight, 
\tc{w,v,z} has to be $m$-tight. 

We modify the four ternary constraints to be not properly $m$-tight and 
modify the four binary constraints \bc{w,u}, \bc{u,z}, \bc{z,v} and \bc{v,w} 
to be properly $m$-tight. 
This preserves the weak $m$-tightness of \networkn{1} and 
the number of properly $m$-tight constraints in \networkn{1}. 

Case 2: Two properly $m$-tight constraints \tc{u,v,w}, and \tc{w,x,y} 
      share one variable $w$. 
      Since \bc{u,w} and \bc{w,x} are not properly $m$-tight, \tc{u,w,x} has 
      to be properly $m$-tight. Since \bc{v,w} and \bc{w,y} are not properly 
      $m$-tight, 
      \tc{v,w,y} has to be properly $m$-tight. Similarly, \tc{u,w,y} and 
      \tc{v,w,x} 
      have to be properly $m$-tight. Now, if we modify the 
      four binary constraints \bc{u,w}, \bc{w,x}, \bc{v,w}, and 
      \bc{w,y} to be properly $m$-tight and the six ternary constraints 
      to be non-properly-$m$-tight, the new network is still weakly $m$-tight
      with fewer $m$-tight constraints.
      This contradicts the minimality of the number of properly $m$-tight 
      constraints in \networkn{1}. Hence, case 2 is not possible.

Step 3. As a result of the first two steps, in 
the network \networkn{1}, the scopes of the properly $m$-tight 
ternary constraints are disjoint, and the binary constraint between 
any two variables of a properly $m$-tight ternary 
constraint is not properly $m$-tight. 

Let $B$ (and $T$ respectively) be the set of the properly $m$-tight binary
(and ternary respectively) constraints of \networkn{1}.

Assume $|T|=k$. Since it is difficult to
count $B$, we count the maximum number of non-properly-$m$-tight 
binary constraints in \networkn{1}.
We have $3k$ non-properly-$m$-tight binary constraints  
due to $T$. We should not have any non-properly-$m$-tight binary 
constraints between a variable in $T$ and a
variable outside $T$. Let $V'$ be the variables outside $T$.
We have $|V'|=n-3k$.
The other non-properly-$m$-tight constraints fall only 
between variables in $V'$.  
Since \networkn{1} is weakly tight at level 2, there is no 
two non-properly-$m$-tight constraints on any 
variable in $V'$. Hence, there are at most $(n-3k)/2$ non-properly-$m$-tight 
constraints if $n-3k$ is even, and otherwise at most $(n-3k-1)/2$ ones.
So the number, denoted by $\delta$, of the properly $m$-tight constraints
in \networkn{1} would be the sum of the cardinality of $T$ and $B$:

\begin{center}
$ \delta = k + (n(n-1)/2 - 3k - \lfloor(n-3k)/2\rfloor)
  = n(n-1)/2 - 2k -  \lfloor(n-3k)/2\rfloor.$ 
\end{center}
The fact that $\delta$ is minimal implies that $k$ should be maximized. 
If $n$ is a multiple of $3$, the number of properly $m$-tight constraints
is $n(n-1)/2 - 2n/3$; if $n$ is $1$ more than a multiple of $3$,
the number is $n(n-1)/2 - 2(n-1)/3$; otherwise the number
is $(n-1)(3n-1)/6$.
\QED

This result shows that under the concept of $k$-consistency 
we still need a significant number of constraints to be properly
$m$-tight to predict the global consistency of a network in terms
of constraint tightness.

\subsection{Dually Adaptive Consistency}
\label{sec:duallyAdaptive}

A main purpose of our characterization of weak $m$-tightness 
of a network is to help identify a consistency condition 
under which a solution of a network can be found without
backtracking, i.e., efficiently. We have studied 
constraint tightness under the concept of $k$-consistency in 
the previous subsections. In this subsection, we introduce
dually adaptive consistency to achieve backtrack free search
by taking into account both the tightness of constraints and the 
topological structure of a network. 

The idea of adaptive consistency \cite{DP87} is 
to enforce only the necessary level of 
consistency on each part of a network to ensure global consistency.
It assumes an ordering on the variables. 
For any variable $x$, it only requires that
a consistent instantiation of the {\em relevant} variables {\em before} 
$x$ can be consistently extensible to $x$. 
Other variables do not play any direct role on $x$ and thus 
are ignored when dealing with $x$. 

We first introduce some notations used in adaptive consistency.

\begin{figure}[htb!]
\begin{center}
\input{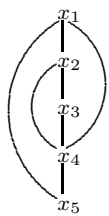}
\end{center}
\caption{\label{fig:width} The variables $\{x_1, x_2, \ldots, x_5\}$
are ordered according to their
subscripts. For example, $x_1$ is before $x_2$. The width of $x_2$ is $1$.
}
\end{figure}

The {\em width} of a variable with respect to a variable ordering
is the number of constraints involving $x$ and only variables before $x$.
See \figref{fig:width} for an example.

Given a network, a variable ordering, and a variable $x$,
the {\em directionally relevant constraints on $x$} are those
involving $x$ and only variables before $x$. In the following,
$DR(x)$ is used to denote the directionally relevant constraints on $x$,
and $S$ used to denote all variables occurring in the constraints of $DR(x)$.

The constraints of $DR(x)$ are {\em consistent} on $x$ if and only if, for
any consistent instantiation $\bar{a}$ of $S-\{x\}$,
there exists $u \in D_x$ such that $(\bar{a},u)$ satisfies all
the constraints of $DR(x)$.

We next define the adaptive consistency of a network. 

\begin{definition} \label{def:adaptiveConsistency}
Given a constraint network and an ordering on its variables, 
the network is {\em adaptively consistent} if and only if for any variable 
$x$, its directionally relevant constraints are consistent on $x$. 
\end{definition}

The adaptive consistency is presented as an algorithm 
by \citeA{Dec03} although, for the purpose of this paper,
we prefer a declarative characterization.

For an adaptively consistent network, a solution can be found without 
backtracking. 

\begin{proposition} \label{prop:adaptiveConsistency}
Given a constraint network and an ordering on its variables, 
a backtrack free search is ensured if the network is adaptively 
consistent.
\end{proposition}
{\bf Proof.} Assume we have found a consistent instantiation of
the first $k$ variables (in terms of the given ordering). 
They can be consistently extended to $x_{k+1}$ because all directionally
relevant constraints on $x_{k+1}$ are consistent on $x_{k+1}$. 
\QED

When a network is not adaptively consistent, the algorithm
by \citeA[p. 105]{Dec03} can be used to enforce adaptive
consistency on it.

Adaptive consistency is not only more accurate
in estimating the local consistency that ensures 
global consistency, but also makes intuitive 
the algorithms to enforce consistency 
and to find a solution.

With the knowledge of constraint tightness 
presented in the previous subsections, we know that 
for a network to be adaptively consistent, 
it is sufficient to make sure that only 
some, not all, directionally relevant constraints on a 
variable are consistent. We are now in a position to
define dually adaptive consistency of a constraint network. 

\begin{definition}
Consider a constraint network and an ordering of its variables.
For any variable $x$ in the network, 
let $c_x$ be one of the tightest directionally relevant constraints
on $x$ and $c_x$ be properly $m_x$-tight.
The network is {\em dually adaptively consistent} if and only if  

1) for any variable $x$ whose width is not greater than $m_x$, 
its directionally relevant constraints are consistent on it, 
and

2) for any variable $x$ whose width is greater than $m_x$,
$c_x$ is consistent with every other $m_x$ directionally 
relevant constraints on $x$. 
\end{definition}

Thanks to the set intersection result of Lemma~\ref{lm:smallSetM}, 
we have the main result on dually adaptive consistency.

\begin{theorem} \label{th:dac}
Given a constraint network and an ordering of its variables, 
a backtrack free search is ensured if it is 
dually adaptively consistent. 
\end{theorem}

\newcommand{\dr}{{\em DR}}

{\bf Proof.}
We only need to prove that the network is adaptively consistent:
For any variable $x$, its directionally relevant constraints \dr$(x)$
are consistent on $x$. Let $S$ be the variables involved in \dr$(x)$.
Consider any consistent instantiation $\bar{a}$ of $S-\{x\}$. We show 
that there exists $u \in D_x$ such that $(\bar{a},u)$ satisfies
constraints in \dr$(x)$. Let $l$ be the number of constraints in \dr$(x)$,
and let $c_x$ be one of the tightest constraint 
in \dr$(x)$ with proper tightness $m_x$.
For any constraint $c_i \in$ \dr$(x)$, 
let $\bar{a}$'s extension set to $x$ under $c_i$ be $E_i$. It is 
sufficient to show
\[
  \cap_{c_i \in \mbox{\dr$(x)$}}E_i \neq \emptyset.
\]
We know $c_x$ is consistent with every other $m_x$ constraints.
Hence, $E_x$, $\bar{a}$'s extension set under $c_x$, intersects with
every other $m_x$ extension sets of $\bar{a}$. 
Lemma~\ref{lm:smallSetM} implies that
\[
  \cap_{c_i \in \mbox{\dr$(x)$}}E_i \neq \emptyset.
\]
\QED

By this theorem,
we need only the tightest of the directionally
relevant constraint on each variable, totally $n-1$ such constraints,
to predict the global consistency of a network. 
This could be considered a significant improvement 
over the results in the previous two subsections. 

Compared with the result by \citeA{DP87}, 
this theorem also provides a lower level (the smaller
of tightness or width) of consistency ensuring 
global consistency. 

When a constraint network is not dually adaptively consistent
with respect to a variable ordering,
it can be made so by enforcing the required consistency
on each variable, in the reverse order of the given variable ordering.
To make the procedure more efficient, we should chose 
a better variable ordering, depending on both the topological
structure of the network and the tightness of the constraints.

\section{Tightness and Convexity Revisited}
\label{sec:set-applicationRC}

The consistency results derived from {\em small set intersection} and {\em
\treeconvex\ set intersection} in
Section~\ref{sec:set-applicationConvexity} and 
Section~\ref{sec:consistencyTightness} can be rephrased in
a relational consistency setting. For example, a new version of weak
tightness based on relational consistency is given as follows.

\begin{theorem}[Weak Tightness] \label{th:weakTightnessRC}
  If a constraint network \network{R} of constraints with arity
  of at most $r$ is strongly relationally $(m+1)$-consistent
  and weakly $m$-tight at level of $(m+1)(r-1)+1$, it is globally consistent.
\end{theorem}

{\bf Proof.} Let $j = (m+1)(r-1)+1$. The constraint network
\network{R} will be shown to be $k$-consistent for all $k$ ($j <
k \le n$).

Let  $Y = \{x_1, \ldots, x_{k-1} \}$ be a set of any $k-1$ variables,
and $\bar {a}$ a consistent instantiation of all variables in
$Y$. Consider any new variable $x_k$. Without loss of generality,
let $c_{S_1}, \ldots, c_{S_l}$ be the relevant constraints on $x_k$, and
$E_i$ the extension set of $\bar {a}$ to $x_k$ with respect to
$c_{S_i}$ for $i \le l$.
\smallskip

({\em Consistency to Set}) Consider any $m+1$ of the $l$ extension
sets. Since \network {R} is relationally $(m+1)$-consistent,
the intersection of $m+1$ extension sets is not empty.
\smallskip

({\em Set to Set}) The network is weakly $m$-tight. So, there must
be a properly $m$-tight constraint in the relevant constraints $c_{S_1},
\ldots, c_{S_l}$. Let it be $c_{S_i}$. Its extension set
$|E_i| \le m$. Since every $m+1$ of the extension sets have a
non-empty intersection, all $l$ extension sets share a common
element by the {\em small set intersection} result
(Corollary~\ref{lm:smallSet}).
\smallskip

({\em Set to Consistency}) From the lifting lemma, we have that
\network{R} is $k$-consistent. \QED

Compared with the weak tightness theorem in 
Section~\ref{sec:consistencyTightness}, the
exposition of the result is neater and the proof is simpler.

For completeness, we also include here a new version
of the \treeconvex\ theorem using relational consistency. 
The proof is omitted since it is a simplified version of the one in
Section~\ref{sec:set-applicationConvexity} as hinted by the proof
above.

\begin{theorem}[Tree Convexity]
Let \network {R} be a \treeconvex\ constraint network. \network{R}
is globally consistent if it is strongly relationally path
consistent.
\end{theorem}
\section{Conclusion}
\label{sec:set-relatedwork}

\hide{
 1. summary of results
 1.1 the framework
 1.2 significant results
 2. application to the work of boi Faltings and P. David (functional)
}

Through the lifting lemma and proof schema, we have shown that
set intersection results can be easily lifted to consistency
results in a constraint network. 
There are a few advantages for this approach of studying
consistency. 

Firstly, although this approach does not offer 
a ``completely new'' way to prove consistency results, it does 
provide a uniform way to understand many seemingly different 
results on the impact of convexity and tightness on global consistency. 
In addition to the results shown here, some
other results can also be obtained easily by the lifting lemma and
proof schema. For example, the  work by \citeA{Dav93} 
can be obtained by lifting the corollary of Lemma~\ref{lm:smallSetM}
\cite{ZY03b}. 
The work by \citeA{SHF96}
on convex constraint networks with continuous domains 
can be lifted from Helly's
theorem \cite{Helly} on the intersection of convex sets in Euclidean spaces.

Secondly, the establishment of the relationship between
set intersection and consistency in a constraint network
makes it easier to communicate the consistency results 
to the researchers outside the constraint network community.
It is also made possible for them 
to contribute to consistency results by exploiting their
knowledge on set intersection properties. 

More importantly, this approach singles out the fact that 
set intersection properties play a fundamental role in 
determining the consistency of a constraint network. 
This perspective helps us focus on properties of set intersection
and discover or generalize the intersection properties of tree convex 
sets and sets with cardinality restrictions. 
The corresponding consistency results have extended our 
understanding of the convexity and tightness of constraints since
Dechter and van Beek's work \citeyear{vBD95,vBD97}. 
We identify a new class of tree convex constraints for which
global consistency is ensured by a certain level of local consistency.
This generalizes row convex constraints by \citeA{vBD95}.
We also show that a weakly $m$-tight constraint network can be
made globally consistent by enforcing local consistency. This type 
of result on tightness is new. Detailed 
study has been carried out on when a constraint network is 
weakly $m$-tight. To make full use of the tightness of the constraints,
we propose dually adaptive consistency that exploits both
the topology and the semantics of a constraint network, which 
again results from the relation between set intersection and consistency. 
Under dually adaptive consistency, the topology of a network and the
tightest relevant constraint on each variable determine the local consistency
that ensures backtrack-free search.

\section*{Acknowledgments}
We are indebted to Dr. Peter van Beek and Dr. Fengming Dong 
for very helpful discussions. The constructive comments from
the anonymous referees of various versions of this paper
have improved its quality.
This material is based on works partially supported by a grant under the
Academic Research Fund of National University of Singapore and by
Science Foundation Ireland under Grant 00/PI.1/C075.
Some materials of this paper appeared in the Proceedings of the 
International Joint Conference on Artificial Intelligence 2003 \cite{ZY03b}
and the Proceedings of Principles and Practice of Constraint Programming 
2004 \cite{Zha04a}.
 
\bibliographystyle{theapa}
\bibliography{cp}

\end{document}